\RequirePackage{silence}
\documentclass[pdflatex,sn-mathphys-num]{sn-jnl}

\usepackage{silence}
\usepackage{amsmath}
\usepackage{bbding}
\usepackage{booktabs}
\usepackage{textcomp} 
\usepackage{gensymb}
\usepackage{graphicx}
\usepackage{lmodern}
\usepackage{makecell}
\usepackage{multirow}
\usepackage{placeins}
\usepackage{subcaption}

\usepackage{lineno}

\usepackage{natbib}
\setcitestyle{super,open=,close=}

\usepackage{siunitx}
\DeclareSIUnit\year{yr}
\DeclareSIUnit\minute{min}

\usepackage{xcolor}
\newcommand{\yy}{\color{green}{\CheckmarkBold}}
\newcommand{\nn}{\color{red}{\XSolidBrush}}

\newsavebox{\measurebox}

\usepackage[shortlabels]{enumitem}

\begin{document}

\title{On the Foundations of Earth and Climate Foundation Models}

\author*[1,2]{\fnm{Xiao Xiang} \sur{Zhu}}\email{xiaoxiang.zhu@tum.de}
\author[1]{\fnm{Zhitong}\sur{Xiong}}\equalcont{These authors contributed equally to this work.}
\author[1]{\fnm{Yi}\sur{Wang}}\equalcont{These authors contributed equally to this work.}
\author[1]{\fnm{Adam J.}\sur{Stewart}}\equalcont{These authors contributed equally to this work.}
\author[1]{\fnm{Konrad}\sur{Heidler}}\equalcont{These authors contributed equally to this work.}
\author[1]{\fnm{Yuanyuan}\sur{Wang}}
\author[1]{\fnm{Zhenghang}\sur{Yuan}}
\author[1]{\fnm{Thomas}\sur{Dujardin}}
\author[1]{\fnm{Qingsong}\sur{Xu}}
\author[3]{\fnm{Yilei}\sur{Shi}}

\affil[1]{\orgdiv{Chair of Data Science in Earth Observation}, \orgname{Technical University of Munich}, \orgaddress{\street{Arcisstraße 21}, \postcode{80333} \city{Munich}, \country{Germany}}}

\affil[2]{\orgname{Munich Center for Machine Learning},  \postcode{80333} \city{Munich}, \country{Germany}}

\affil[3]{\orgdiv{School of Engineering and Design}, \orgname{Technical University of Munich}, \orgaddress{\street{Arcisstraße 21}, \postcode{80333} \city{Munich}, \country{Germany}}}

\abstract{Foundation models have enormous potential in advancing Earth and climate sciences, however, current approaches may not be optimal as they focus on a few basic features of a desirable Earth and climate foundation model. Crafting the ideal Earth foundation model, we define eleven features which would allow such a foundation model to be beneficial for any geoscientific downstream application in an environmental- and human-centric manner. We further shed light on the way forward to achieve the ideal model and to evaluate Earth foundation models. What comes after foundation models? Energy efficient adaptation, adversarial defenses, and interpretability are among the emerging directions.}


\maketitle


As deep learning continues to mature, a general trend towards larger datasets (with millions to billions of data points~\cite{dehghani2023scaling}) and larger model architectures (with billions to trillions of parameters~\cite{hudson2024trillion}) has emerged. In the past decade in particular, we have witnessed a paradigm shift from single-purpose models to general-purpose models, and from supervised pre-training to self-supervised pre-training. This has been led by the rise in popularity of models like CLIP~\cite{radford2021learning}, LLaMA~\cite{touvron2023llama}, GPT~\cite{achiam2023gpt}, and SAM~\cite{kirillov2023segment} spawning billion dollar companies and revolutionizing many aspects of daily life.

\begin{figure}[htbp]
    \centering
    \includegraphics[width=1.0\textwidth]{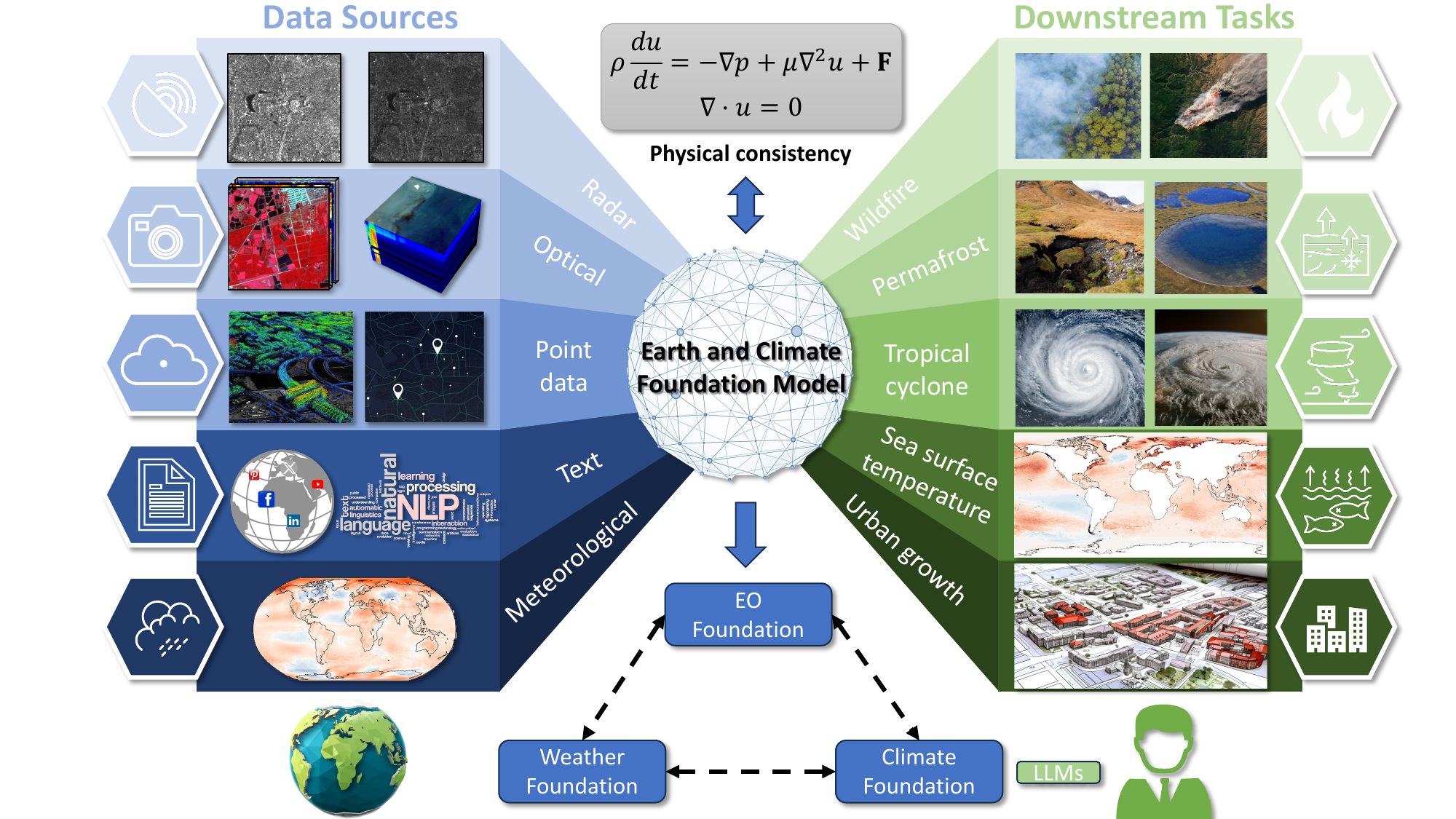}
    \caption{The scheme of an Earth and climate FM. It should be trained on common data modalities, including imagery (radar, optical), non-Euclidean data (point clouds, text), and meteorological data. It should also provide consistency with physical laws. The FM is task-agnostic, exemplified by five possible downstream tasks. Because of the large difference of the characteristics among Earth observation, weather, and climate data, the resulting FM may consists of a pool of expert models, where feedback loops exist among each of them. For attribution of figure elements, please see Supplementary Information.}
    \label{fig:ECFM}
\end{figure}

\citet{bommasani2021opportunities} proposed the term ``foundation model'' (FM) to encompass ``any model that is trained on broad data (generally using self-supervision at scale) that can be adapted (e.g., fine-tuned) to a wide range of downstream tasks'' in order to study this phenomenon. The majority of FMs like CLIP and GPT focus on the image and text domains. In this work, we specifically focus on ``data'' and ``downstream tasks'' relating to the Earth and its climate system, as shown in Fig.~\ref{fig:ECFM}.


We choose to limit the scope of our work to the Earth's surface and atmosphere for three reasons. First, the Earth's surface and troposphere are our home, and include the majority of processes that directly impact and are impacted by human activity. Second, the ``big data'' revolution required to pre-train FMs can currently only be provided at scale using remote sensing data, which is only available at the surface. Third, the timescale at which the deep ocean and mantle operate is too slow to study in combination with a human timescale, and can be considered to be more or less static.

\section*{The Potential of Earth Foundation Models}
Earth and climate science has inevitably entered the era of big data. This data, largely unlabeled and vast in volume, can offer novel insights into our planet's systems. Earth observation (EO) data-informed Earth and Climate modeling offers perspectives towards higher resolution and less biased future projections. However, the sheer volume and redundancy of this information calls for more sophisticated analytical frameworks~\cite{zhu2017deep}. FMs are proving to be a glimmer of hope in this context. This section delves into the critical reasons underpinning the need for FMs in the Earth and climate sciences, illustrating their potential to revolutionize our understanding of the Earth system towards its sustainable management.

\begin{itemize}
\item \textbf{Unlocking the value of big EO and climate data:} The primary allure of FMs lies in their ability to harness the vast, often unlabeled, reservoirs of EO and climate data by means of learning general representations from them~\cite{bommasani2021opportunities}. Traditional models falter under the weight of this information overload, struggling to process and extract meaningful insights without extensive, manually labeled datasets. FMs, on the other hand, can learn task-agnostic general feature representations from massive amounts of unlabelled data. This capability not only improves our understanding of Earth's systems, but also accelerates the pace of discovery from the learned general features.

\item \textbf{Enhancing label efficiency/boosting performance in downstream tasks:} Downstream EO tasks often suffer from a lack of labeled data; a bottleneck that can significantly impact the training and performance of models. FMs address this limitation by being inherently more efficient. They are trained through self-supervised learning on large, unlabeled datasets, which can then be fine-tuned with smaller, task-specific datasets, reducing the reliance on extensive labeled data~\cite{wang2022self1}. This efficiency not only accelerates the applications of artificial intelligence (AI) for EO, but also improves their accuracy, robustness and reliability.

\item \textbf{Reducing carbon footprints:} The environmental impact of training AI models from scratch for every new use case cannot be overstated~\cite{strubell2019energy}. Each training cycle consumes considerable computational resources, contributing to the carbon footprint of AI research and applications. FMs offer a sustainable alternative by allowing for the reuse and adaptation of pre-trained models across multiple use cases. This approach significantly reduces the computational demand and, consequently, the environmental impact of AI model training in the Earth and climate sciences.

\item \textbf{Bridging EO and climate science:} The synergy between EO and climate science is pivotal for the development of informed climate models, in particular, given the severe spatial and time-span mismatch between high resolution observations and climate modeling and projections. Earth and climate FMs facilitate this integration by providing a framework through which EO data can directly inform and enhance climate models. This is essential for improving the resolution and prediction accuracy of climate models, and reducing their biases and uncertainty, and thus leads to more effective and sustainable climate change mitigation and adaptation strategies.

\item \textbf{Improving Earth system modeling:} The study of climate change trends and the prediction of future weather and climate scenarios have traditionally relied on general circulation models and numerical weather prediction models. However, the complexity and scale of Earth system modeling (ESM) pose challenges to computational capacity and efficiency with these conventional methods. FMs offer scale invariance and efficient feature representations, presenting an opportunity for a unified approach to ESM~\cite{chen2023foundation}. Furthermore, ESM's complexity and uncertainties require an interdisciplinary methodology to advance scientific knowledge discovery. Generative FMs show promise in analyzing diverse ESM data modalities, including images and sequences, to facilitate tasks such as physical law discovery~\cite{wang2023scientific,hudson2024trillion}.
\end{itemize}

To this end, advances in Earth and climate FMs would lead to a paradigm shift in how we approach the vast and complex observational and simulated data characterizing our planet. By unlocking the value of big EO and climate data, enhancing label efficiency, reducing carbon footprints, and bridging the gap between EO and climate science, FMs pave the way for a more sustainable, efficient, and comprehensive understanding of the Earth system. Their potential to revolutionize Earth and climate sciences is immense, in particular when it comes to inform critical decisions in the face of climate change, e.g. by improving impact modeling and ESM.

\section*{Big (Earth) Data}\label{sec:data-sources}

\begin{figure}[htbp]
    \centering
    \begin{subfigure}[b]{0.7\textwidth}
        \includegraphics[width=\textwidth]{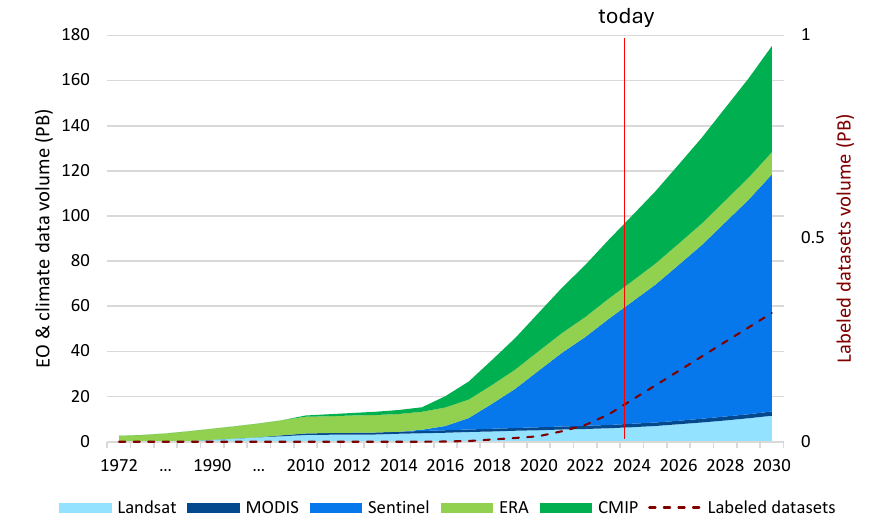}
        \caption{Historical, current, and estimated future volume of open EO and climate data archives in petabytes~\cite{annualreport,crawford202350,soille2018versatile,copernicus2023era5,bell2021era5,petrie2020coordinating,xiong2022earthnets}. Labeled dataset volumes are shown by a dashed line and use a different \(y\)-axis scale. Note that the volume of labeled data is less than 0.1\% of the total volume of unlabeled data. 
        }
        \label{fig:open-eo-volume}
    \end{subfigure}

    \begin{subfigure}[b]{0.7\textwidth}
        \includegraphics[width=\textwidth]{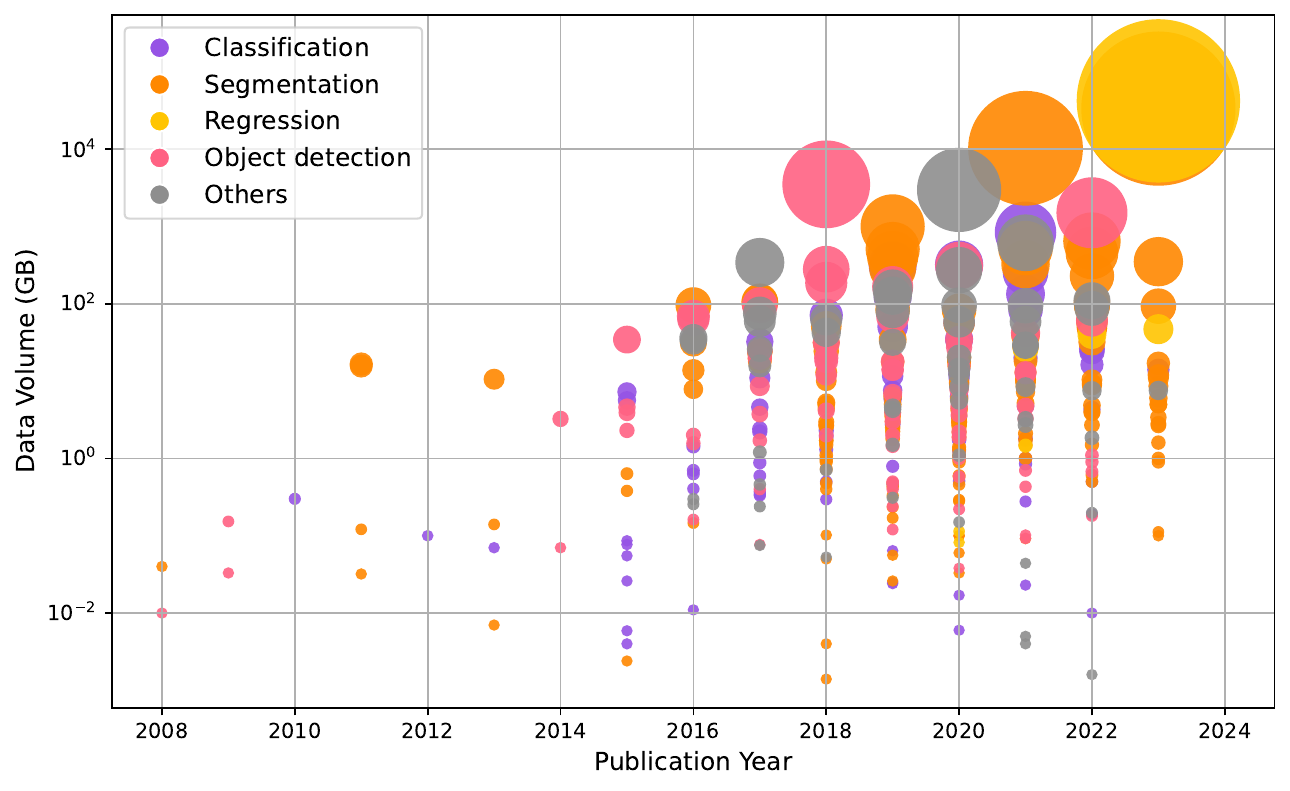}
        \caption{Evolution of labeled EO datasets over the last 16 years~\cite{xiong2022earthnets}. Bubble size and \(y\)-axis height indicate the volume of labeled data in each dataset. Bubble color denotes the type of dataset. 
        }
        \label{fig:benchmark-eo-volume}
    \end{subfigure}
    \caption{Volume of EO and climate data archives and curated benchmark datasets.}
    \label{fig:eo-volume}
\end{figure}

The development of Earth and climate FMs relies on vast and varied data for self-supervised pre-training. Below, we delve into the most important data sources for both training and evaluation.

\begin{itemize}
\item \textbf{EO data archive:} To date, there are more than 1000 active remote sensing satellites in space~\cite{orbitnow}. Various EO archives provide a huge volume of data sources for building EO FMs. As Fig.~\ref{fig:open-eo-volume} shows, the data volume of EO missions whose data is open and free drastically increased over the past decade, surging to nearly 100~PB today, not including commercial data with comparable volume. These archives provide a continuous, high-quality, and open stream of satellite imagery, capturing a wide range of Earth's features at various spatial, spectral, and temporal resolutions. For example, the Sentinel series~\cite{malenovsky2012sentinels} includes multiple satellites with different sensors designed to monitor the land, ocean, and atmosphere, and the Landsat series~\cite{wulder2022fifty} commands attention for its extensive temporal reach, providing an invaluable long-term perspective on Earth's surface changes.


\item \textbf{Weather patterns and climate models:} In meteorology, various weather and atmospheric science agencies, including ECMWF and NOAA, provide accurate and comprehensive datasets for understanding the Earth's climate system. Reanalysis products such as ERA5~\cite{hersbach2020era5}, JRA-55~\cite{kobayashi2015jra}, NCEP~\cite{kalnay2018ncep}, and HRRR~\cite{dowell2022high} are already widely used for FM pre-training. High-resolution, historical, and near-real-time data on atmospheric conditions make it possible to build large-scale weather nowcasting and medium-range forecasting systems. Furthermore, climate models like CMIP6~\cite{o2016scenario} and MERRA-2~\cite{gelaro2017modern} offer simulations of past, present, and future climate conditions built on strong physical understandings of the climate system that are valuable to climate FMs. Fig.~\ref{fig:open-eo-volume} highlights a number of these datasets.

\item \textbf{Curated ML-ready datasets:} The rise of machine learning (ML) in the past decade has driven the EO community to curate and publish over 500 ML-ready datasets~\cite{xiong2022earthnets}. Integrated with human annotation, these supervised datasets cover a wide range of applications, such as land-use land-cover mapping~\cite{sumbul2019bigearthnet}, change detection~\cite{toker2022dynamicearthnet}, object detection~\cite{xia2018dota}, and disaster monitoring~\cite{xu2023disasternets}, as illustrated in Fig.~\ref{fig:benchmark-eo-volume}. The diverse array of applications along with semantic labels can steer the FMs toward nuanced understanding and broad applicability. Several recent works on FMs have also released a set of curated pre-training datasets without labels~\cite{manas2021seasonal,wang2023ssl4eo,stewart2024ssl4eo,bastani2023satlaspretrain}. Compared to the raw EO archive, these datasets are compact yet comprehensive, crafted with strategic sampling strategies that enhance the efficiency of model training. Nevertheless, in contrast to EO data archive, existing labelled datasets in EO amounts to less than 0.1~PB to date (see the right \(y\)-axis of Fig.~\ref{fig:open-eo-volume}), i.e., only about 0.1\% of the archive of open EO data, highlighting the huge discrepancy between them.
\end{itemize}
Apart from the above, other sources such as natural language texts, cadastral records, and historical in-situ measurements are also valuable for the training of Earth and climate FMs.

\section*{Crafting the Ideal Earth and Climate Foundation Model}
\label{sec:idealFM}
An ideal Earth and climate FM (see Fig.~\ref{fig:ECFM}) should take domain-specific characteristics in Earth and climate sciences into account, and thus \textbf{must} feature:
\begin{enumerate}
\item \textbf{Geolocation embedding:} EO and climate data are linked to a geolocation. Embedding geolocation into the FM, either in a hard-coded way by importing the latitude and longitude as inputs, or in a smoother manner by contrasting features according to their geographical distances, can improve feature representation or even facilitate the study of localized events, such as heatwaves or hurricanes. Researchers can potentially pinpoint the exact locations of environmental phenomena, enabling precise analysis of regional climate patterns and their global implications. 

\item \textbf{Balanced geographical representations:} To date, ML models for Earth and climate sciences have been trained with a bias towards data-rich regions. To take urban applications as an example, it is a common practice to consider open source OpenStreetMap~\cite{OpenStreetMap} data for training models, leading to a severe bias in favor of Europe and North America. A balanced geographical coverage implies a careful selection of the data used for training the FMs, to ensure that the FMs accurately reflect the diversity of Earth’s surfaces, climates, and ecosystems.

\item \textbf{Scale awareness:} The spatial resolution of EO and climate data varies from sub-meter for high-resolution aerial images to hundreds of kilometers for global climate models. The ideal FM should have scale awareness of different input data and learn spatial features referring to a certain characteristic scale. Practically, this is a mandatory feature for the model to deal with scale variances of the input data. More importantly, the FM should be capable of examining large-scale patterns, such as ocean circulations, while also delving into finer-scale processes, like land cover changes.

\item \textbf{Wavelength embedding:} Every individual satellite has a unique specification in terms of wavelengths and number of bands. The ideal FM must incorporate data across different wavelengths, including visible light, infrared, ultraviolet, and microwave, and possess different numbers of channels, ranging from one (e.g., panchromatic) to several hundred (e.g., hyperspectral). Fine-tuning pre-trained models should be possible with input data of any wavelength and number of bands, including completely new and unseen sensors.

\item \textbf{The time variable:} EO and climate time-series data offer valuable new capabilities in monitoring the changing planet and making predictions with unprecedented spatial and temporal resolution. Therefore, the ideal FM must be able to learn sequential relations of data within its spatial context. This feature is vital, e.g., for distinguishing between natural variability and anthropogenic impact on the Earth surface, detecting anomalies, and identifying emerging patterns. 

\item \textbf{Multisensory:} Multi-source and multisensory EO and climate data offer complementary information in order to serve various tasks. The FM must be capable of learning general feature representations across all data sources while retaining the unique characteristics of each source. This dual capability ensures that the model can draw upon a broad range of data and explore their joint and source-specific features to build a comprehensive picture of the Earth and its climate system. Meanwhile, preserving source-specific features allows for, e.g., the accurate attribution of climate phenomena to their respective causes.

\item \textbf{Task-agnostic:} In Earth and climate sciences, AI models have been trained to serve various tasks, such as image classification, semantic segmentation, detection, regression, and forecasting. As a FM, it must be task-agnostic yet beneficial for a wide spectrum of downstream use cases. Potential benefits are, for example, enhanced label efficiency, boosted performance, improved generalizability and transferability, and better robustness against label noise.

\item \textbf{Carbon minimized:} Given the urgency of addressing climate change, it is indispensable that the development and operation of FMs minimize their carbon footprint. This can be achieved through unified models for more general purposes, efficient computing practices, the use of renewable energy sources, and the optimization of model architectures. By prioritizing sustainability in its design, the FM aligns with the broader goals of reducing greenhouse gas emissions and promoting environmental stewardship.
\end{enumerate}

In addition to the above features that are considered must-haves, we also consider a number of features that are \textbf{highly desirable} for the ideal Earth and climate FM. 
\begin{enumerate}[resume*]
\item  \textbf{Uncertainty quantification:} Quantifying uncertainty and detecting out-of-distribution instances is a crucial ability for FMs, especially when adapted to extreme Earth events. In this situation, data scarcity and unexpected situations can significantly impact the performance and robustness of downstream tasks. Considering this, enabling FMs with uncertainty quantification and out-of-distribution detection ability can make them more reliable in critical applications, such as disaster response.

\item \textbf{Physical consistency:}
Purely data-driven FMs can make predictions based on non-physical relationships or spurious correlations~\cite{chen2023foundation}. Incorporating principles of physical consistency, such as conservation, symmetry, and causality, improves the pre-training and fine-tuning of FMs, enhancing their transferability to new domains and increasing their transparency. 
Physically consistent models, alternatively termed knowledge-guided or physics-aware ML~\cite{xu2023physics, kashinath2021physics}, represent a promising research direction to address physical consistency of Earth and climate FMs.

\item \textbf{AI assistants:} Developing ML systems based on the Earth and climate FMs to understand and interpret complex EO data in a human-like manner would provide researchers and decision-makers with intuitive insights and recommendations tailored to specific scenarios. This approach can further enrich common user interactions through interfaces with the incorporation of natural language, making the model more user-friendly and facilitating the communication of EO and climate science to a wider audience.
\end{enumerate}

\begin{figure}[htbp]
    \centering
    \includegraphics[width=\textwidth]{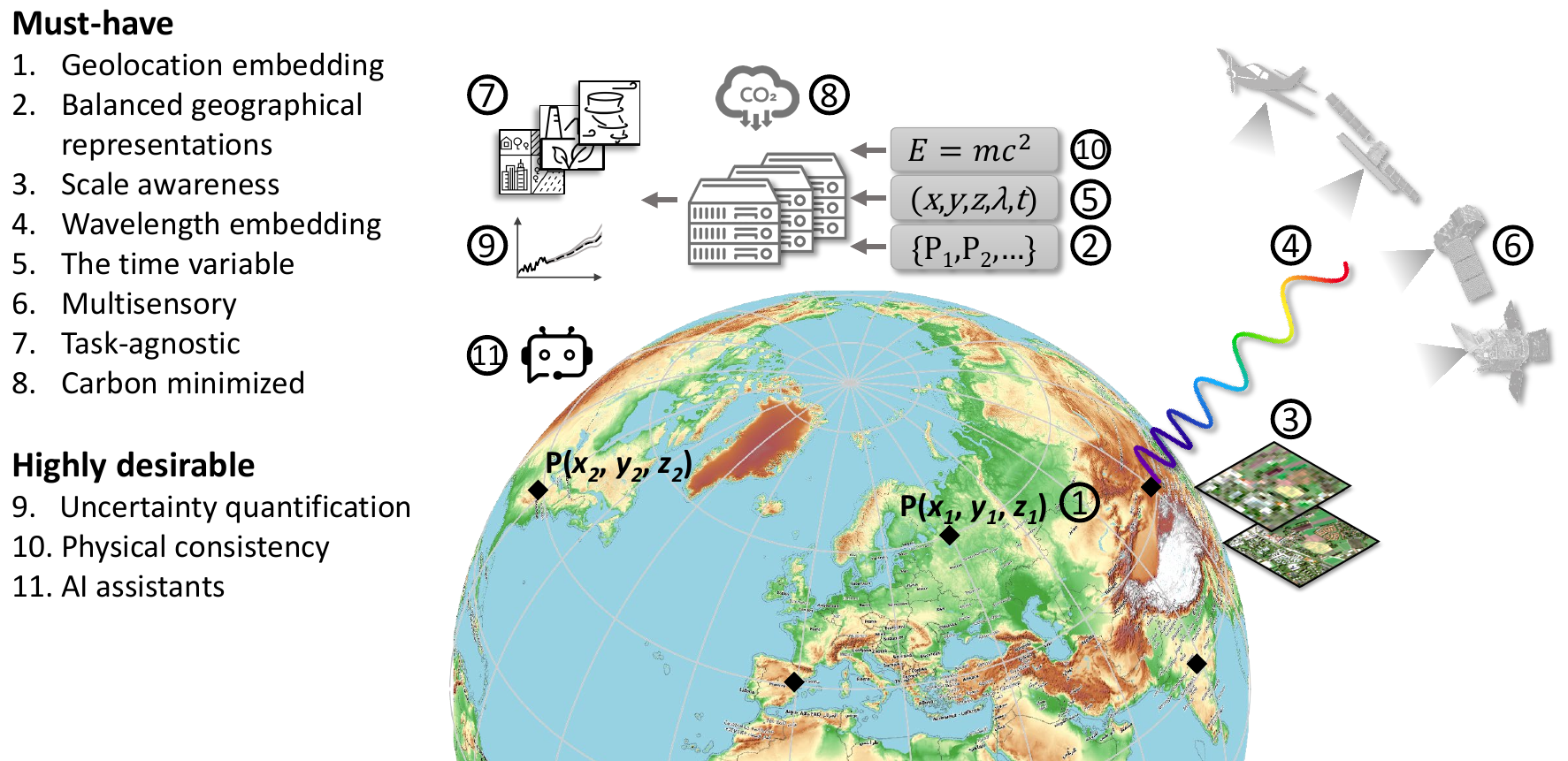}
    \caption{The ideal Earth and climate FM. It should possess at least eight ``must-have'' features and three ``highly desirable'' features. Among them, the must-haves (1--8) include geolocation embedding, balanced geographic representations, scale awareness, wavelength embeddings, the time variable, multisensory, task-agnostic and carbon minimized. They define the basic functionality of an ideal Earth and climate FM, i.e., the ability to support any downstream task regardless of input data at any location in an environmentally friendly manner. The highly desirable features (9--11) are uncertainty quantification, physical consistency, and ML assistants, ensuring the trustworthiness and human-centric design of an ideal Earth and climate FM. For attributions of figure elements, please see Supplementary Information.}
    \label{fig:perspective_FM}
\end{figure}

\section*{Recent Advances and Gaps}

\begin{figure}[htbp]
    \centering
    \sbox{\measurebox}{
        \begin{minipage}[b]{0.55\textwidth}
        \subfloat[Comparison between EO and climate spatial resolutions.]{
            \includegraphics[height=6cm]{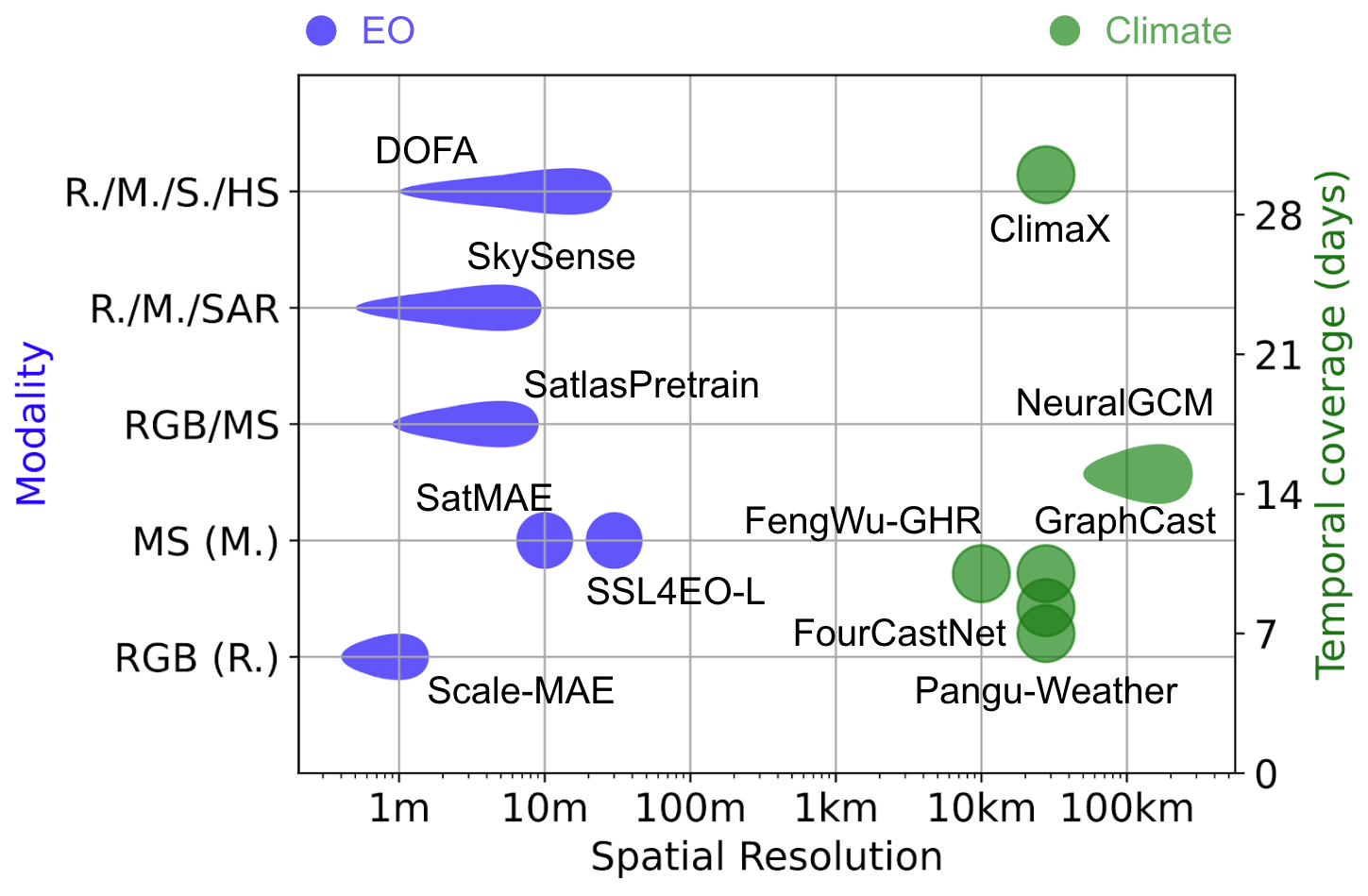}
            \label{fig:eo_cl_fm}
        }
        \end{minipage}
    }
    \usebox{\measurebox}\hfill
    \begin{minipage}[b][\ht\measurebox][s]{0.43\textwidth}
        \centering
        \subfloat[EO model comparison.]{
            \includegraphics[height=2.5cm]{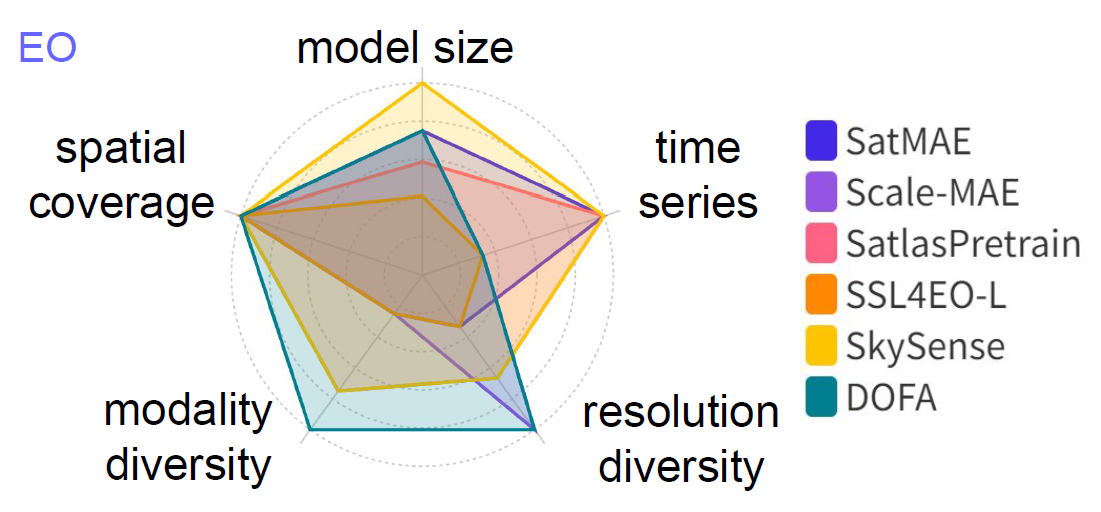}
            \label{fig:eo_fm}
        }
        \vfill
        \subfloat[Climate model comparison.]{
            \includegraphics[height=2.5cm]{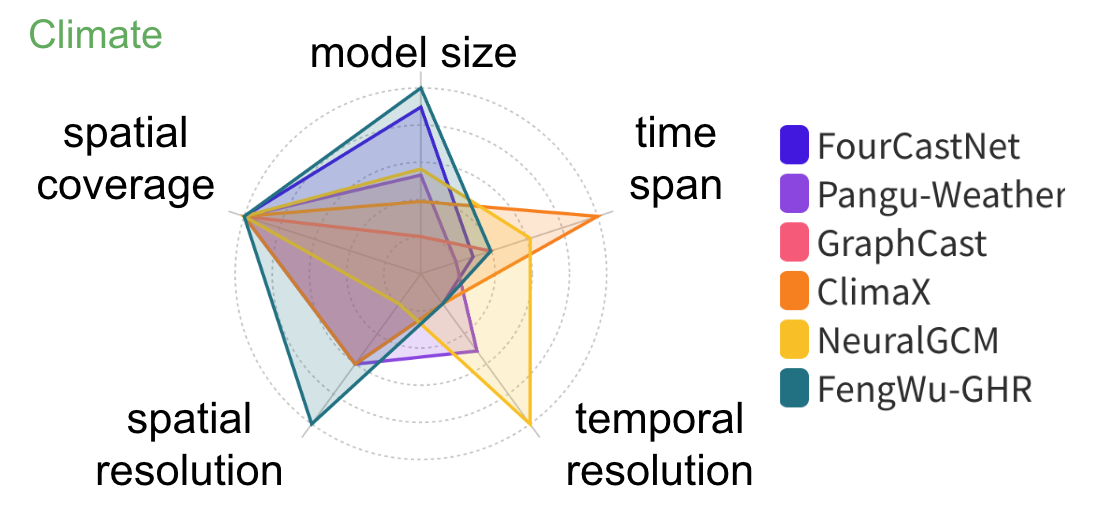}
            \label{fig:cl_fm}
        }
    \end{minipage}
    \caption{Representative EO and climate FMs. Representative models are chosen based on both popularity and novelty. Note that no single model excels at all evaluation criteria, as most models focus on only one or two of all ``must-have'' features.}
    \label{fig:eo_cl_fm_combined}
\end{figure}

Pioneering studies has been conducted on the development of Earth and climate FMs. However, compared to the ideal FM we describe above, existing models are still in their infancy, focusing on two or three of the above ``must-have'' features but rarely approaching coverage of all eight or addressing any ``highly desirable'' features. In this section, we discuss the state of the art and current limitations with a focus on EO and climate FMs, which are typically developed separately and with different downstream applications in mind. A general view of several representative EO and climate models is shown in Fig.~\ref{fig:eo_cl_fm}, with the spatial scale potentially linking the two fields in monitoring the same planet. As it stands, there exists a clear gap in the spatial resolution at which EO and climate models operate. In order to create a unified EO and climate model, not only does the spatial connection need to be consolidated, but also other aspects such as the time dimension need to be considered.

\subsection*{Earth observation foundation models}

EO has a relatively rich record of FM development, owing in part to the many similarities between remotely-sensed and natural images and the abundance of unlabeled data (Fig.~\ref{fig:open-eo-volume}). Existing architectures, self-supervised techniques, and data augmentation methods can be conveniently adopted, leading to the widespread availability of EO FMs~\cite{wang2022self}. Similar to computer vision models, EO models demonstrate \emph{task-agnostic} representation capabilities when trained on million image-scale datasets~\cite{bastani2023satlaspretrain,guo2023skysense}. Nevertheless, the unique characteristics of EO data and applications call for domain-specific considerations~\cite{rolf2024mission}, leading to many exciting recent advances.

Many researchers start with careful curation of pre-training datasets in order to achieve \emph{geographical balance}. SeCo~\cite{manas2021seasonal} and SSL4EO~\cite{wang2023ssl4eo,stewart2024ssl4eo} rely on sampling from regions with high spatial variability, while others like SatlasPretrain~\cite{bastani2023satlaspretrain} focus on scale, with imagery covering a total area of 21~million~km\textsuperscript{2}, roughly 10\% of the total land surface. Other researchers focus on developing novel self-supervised techniques designed to incorporate \emph{geolocation embedding}, including Tile2Vec~\cite{jean2019tile2vec} via contrastive learning and GASSL~\cite{ayush2021geography} via an explicit pretext task. Scale-MAE~\cite{reed2023scale} uses the ground sampling distance as input to the ViT positional encoder, allowing for explicit \emph{scale awareness}. A number of recent FMs, including SatMAE~\cite{cong2022satmae}, Presto~\cite{tseng2023lightweight}, Prithvi~\cite{jakubik2023foundation}, SatlasPretrain~\cite{bastani2023satlaspretrain}, and SkySense~\cite{guo2023skysense}, also incorporate \emph{time}. 

Support for \emph{multisensor} inputs has also required a number of architectural and optimization modifications~\cite{fuller2024croma,guo2023skysense,wang2023decur,xiong2024one,xiong2024neural}, owing to differences between the number of spectral bands and wavelengths captured by each sensor. Among them, DOFA~\cite{xiong2024neural} currently supports the widest range of sensor modalities, including SAR, RGB, MSI, and HSI. It incorporates \textit{wavelength embedding} using a wavelength-based dynamic weight generation strategy, allowing it to adapt to unseen spectral signals. A number of recent papers explore integration between vision and language models in EO~\cite{liu2023remoteclip,wang2023skyscript,zeng2023global,zhang2024earthgpt}, paving the way for \emph{AI assistants}.

\subsection*{Weather and climate foundation models}

Compared to EO, the use of ML for weather forecasting and climate modeling is relatively recent. One of the greatest successes of climate FMs has been \emph{minimizing the carbon footprint} of weather forecasting. As an example, FourCastNet~\cite{pathak2022fourcastnet} is capable of generating a week-long global forecast in less than two seconds on a single GPU, requiring 12,000 times less energy than ECMWF's IFS~\cite{wedi2015modelling}. Faster prediction times also enable larger ensembles, bolstering capabilities for \emph{uncertainty quantification}.

The majority of modern weather and climate FMs are trained on \emph{multisensor} reanalysis products like ERA5 at a global scale, and thus innately offer \emph{geolocation embedding} and \emph{balanced geographical representations}. Although ERA5 limits many models to a spatial resolution of 0.25\degree{}, a notable exception is the recent FengWu-GHR~\cite{han2024fengwu} model, which brings the spatial resolution down to 0.09\degree{} (about 10km) using priors from pre-trained high resolution models.

Weather and climate forecasting also happen at a number of \emph{temporal} scales, including nowcasting (0--2 hours), medium-range forecasting (2 hours--2 weeks), subseasonal (2 weeks--2 months) to seasonal (2 months--2 years) forecasting, and true ``climate'' modeling (decadal scale). The majority of weather FMs fall under ``medium-range forecasting'', and predict a number of \emph{task-agnostic} atmospheric and surface phenomena every 6 hours. The first weather FM to outperform traditional numerical models, Pangu-Weather~\cite{bi2022pangu}, combines multi-resolution models to allow for predictions every 1 hour. Subseasonal models like DLWP~\cite{weyn2021sub} and true ``climate'' models like ACE~\cite{watt2023ace} are less common, and even fewer models are able to handle multiple temporal scales like ClimaX~\cite{nguyen2023climax}.

In terms of \emph{physical consistency}, NeuralGCM represents a unique and promising modeling direction, with a fully differentiable general circulation model encompassing many physics-based atmospheric dynamics. The rapid progress of large language models has also impacted climate models, making them more user-friendly and widely applicable through natural language integration. For instance, ClimateBert~\cite{webersinke2021climatebert} facilitates climate text analysis, and ClimateGPT~\cite{thulke2024climategpt} serves as an \emph{AI assistant} for climate change and sustainability topics.

\subsection*{Gaps towards the ideal model}

Despite significant advances towards many of the ``must-have'' and ``highly desirable'' features of an ideal FM, large gaps still remain. Works like SeCo and SSL4EO that focus on pre-training dataset curation tend to oversample from densely-populated regions like cities and undersample from underpopulated regions like rainforests, polar regions, and oceans, each critical for ecology, climatology, and oceanography. Furthermore, pre-training datasets tend to focus on medium- to high-resolution RGB and MSI, with few large-scale datasets for SAR, HSI, or Lidar data or for very high- or low- resolution imagery~\cite{wang2022self}. Multisensor models like DOFA are rare, and most models need to be trained from scratch on every new imaging platform encountered, making it difficult to take advantage of the thousands of satellites in orbit. Time-series EO modeling is also in its infancy, with few models able to handle variable length image sequences with irregular spacing. Integrating uncertainty quantification and physical consistency in EO remains underexplored.

Weather and climate FMs have come a long way in a short period of time, yet still show similar gaps in many aspects. Although inference is several orders of magnitude faster than traditional numerical models, training can be a bottleneck for many researchers. For example, NeuralGCM required 10 days to train on 128 TPU v5e, equivalent to roughly 130K hours on a single NVIDIA A100 GPU. Most FMs are currently limited by the 0.25\degree{} resolution of ERA5, and struggle to approach higher resolutions. Medium-range forecasting has seen great progress, while nowcasting, subseasonal to seasonal (S2S) forecasting, and climate projection remain underexplored. Weather FMs suffers from overly smooth forecasts and bias drift as lead time increases~\cite{ben2023rise}, making long-range forecasting an active challenge. 

As seen in Fig.~\ref{fig:eo_cl_fm}, there remains a large gap in spatial resolution between the EO and climate domains, despite obvious connections between surface processes and weather. Table~\ref{tab:gap_fm} illustrates current progress towards the ideal FM with several prominent examples.


\begin{table*}[htbp]
\small
\newcommand\dd{90}
\centering
\begin{tabular}{cccccccccccc}
\textbf{FMs} & \rotatebox{\dd}{\textbf{1. Geolocation Embedding}} & \rotatebox{\dd}{\textbf{2. Balanced Distribution}} & \rotatebox{\dd}{\textbf{3. Scale Awareness}} & \rotatebox{\dd}{\textbf{4. Wavelength Embedding}} & \rotatebox{\dd}{\textbf{5. The Time variable}} & \rotatebox{\dd}{\textbf{6. Multisensory}} & \rotatebox{\dd}{\textbf{7. Task-agnostic}} & \rotatebox{\dd}{\textbf{8. Carbon-minimized}} & \rotatebox{\dd}{\textbf{9. Uncertainty}} & \rotatebox{\dd}{\textbf{10. Physical Consistency}} & \rotatebox{\dd}{\textbf{11. Language Alignment}} \\ \midrule
SatMAE~\cite{cong2022satmae} & \nn & \nn & \nn & \yy & \yy & \nn & \yy & \nn & \nn & \nn & \nn \\ 
Scale-MAE~\cite{reed2023scale} & \nn & \nn & \yy & \nn & \nn & \nn & \yy & \nn & \nn & \nn & \nn \\ 
SatlasPretrain~\cite{bastani2023satlaspretrain} & \nn & \nn & \nn & \nn & \yy & \yy & \yy & \nn & \nn & \nn & \nn \\ 
SSL4EO-L~\cite{stewart2024ssl4eo} & \nn & \yy & \nn & \nn & \nn & \yy & \yy & \nn & \nn & \nn & \nn \\ 
SkySense~\cite{guo2023skysense} & \yy & \nn & \nn & \nn & \yy & \yy & \yy & \nn & \nn & \nn & \nn \\ 
DOFA~\cite{xiong2024neural} & \nn & \nn & \nn & \yy & \nn & \yy & \yy & \yy & \nn & \nn & \nn \\ 
\midrule
FourCastNet~\cite{pathak2022fourcastnet} & \yy & \yy & \nn & - & \yy & \yy & \yy & \yy & \yy & \nn & \nn \\ 
Pangu-Weather~\cite{bi2023accurate} & \yy & \yy & \nn & - & \yy & \yy & \yy & \yy & \yy & \nn & \nn \\ 
GraphCast~\cite{lam2023learning} & \yy & \yy & \nn & - & \yy & \yy & \yy & \yy & \nn & \nn & \nn \\ 
ClimaX~\cite{nguyen2023climax} & \yy & \yy & \nn & - & \yy & \yy & \yy & \nn & \nn & \nn & \nn \\ 
NeuralGCM~\cite{kochkov2023neural} & \yy & \yy & \yy & - & \yy & \yy & \yy & \nn & \nn & \yy & \nn \\ 
FengWu-GHR~\cite{han2024fengwu} & \yy & \yy & \yy & - & \yy & \yy & \yy & \nn & \nn & \nn & \nn \\ 
\bottomrule
\end{tabular}
\caption{Representative EO and climate FMs and which of the 11 features of an ideal FM have been addressed in their model design or training procedure. Wavelength embedding is ignored for climate FMs due to the lack of a wavelength component in atmospheric measurement data. Note that many climate FMs are carbon-minimized at inference time (compared to numerical weather prediction models), but still require substantial computing resources during training.}
\label{tab:gap_fm}
\end{table*}
\section*{The Way Forward}

\begin{figure}[htbp]
    \centering
    \includegraphics[width=0.9\textwidth]{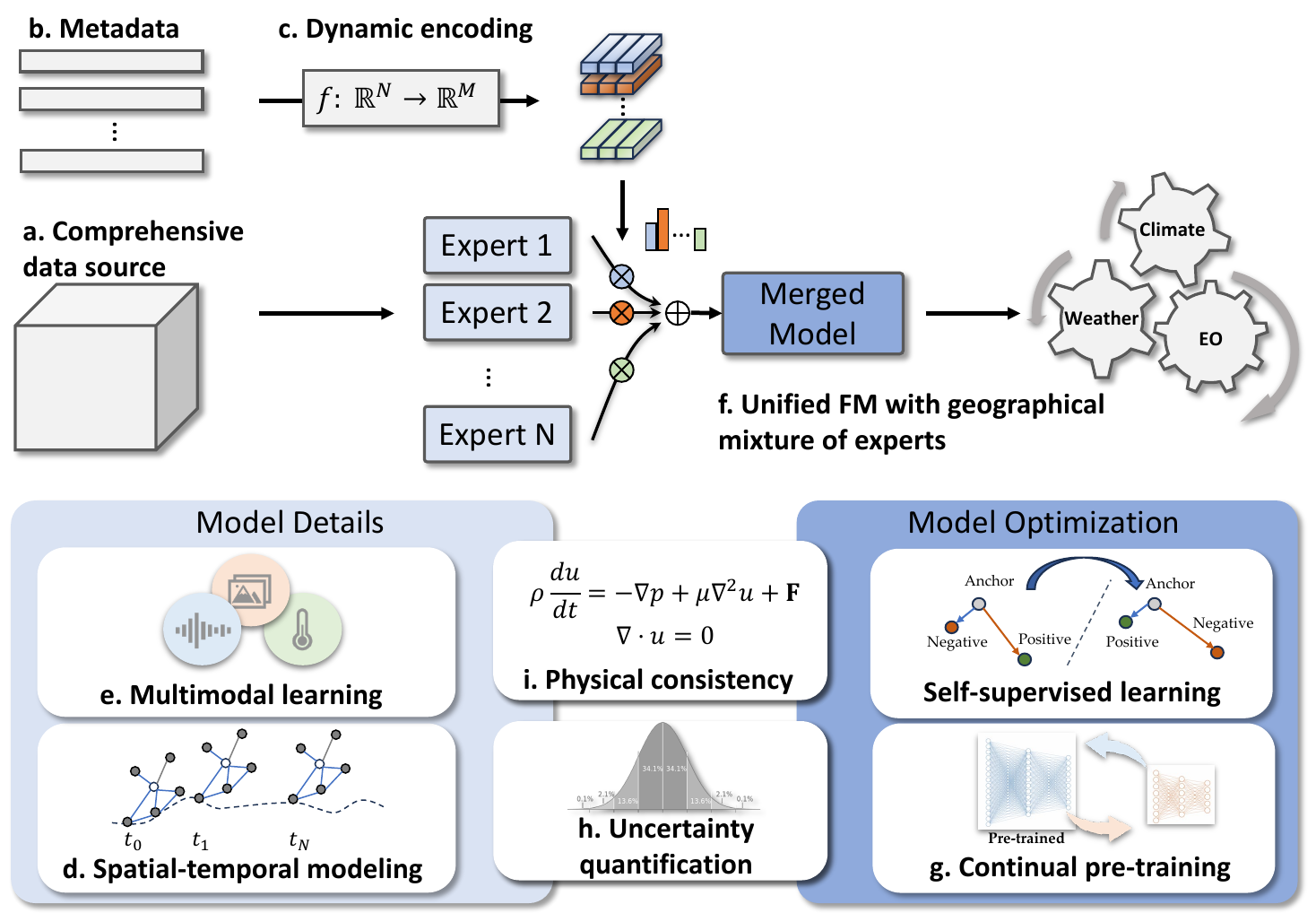}
    \caption{The way forward towards the ideal Earth and climate FM, from data curation to model design and training. We illustrate the design principles to achieve a unified FM, including dynamic encoding, spatial-temporal modeling, geographical mixture of experts, task-agnostic multimodal learning, carbon minimization, physical consistency, uncertainty quantification, and language alignment.}
    \label{fig:ways2ideal}
\end{figure}

To achieve the ideal Earth and climate FMs, two factors should be carefully considered: 1) design a comprehensive data construction process to ensure the high-quality of training data; 2) develop model architectures with key features tailored to the specific challenges of these domains. Fig.~\ref{fig:ways2ideal} presents a general workflow of different specific modules, which we will dig into in the following.


\subsection*{Data curation}
\begin{enumerate}[a.]
    \item \textbf{Comprehensive data sources}: Diversity is the foundation of an effective large-scale pretraining dataset. Satellite imagery, ground-based sensors, weather stations, and climate models are all important sources of data for model training. The diverse data sources ensure a holistic representation of the Earth's system, and capture phenomena across scales and conditions. To ensure geographical diversity, it is vital to include data across evenly distributed geographical regions to avoid data bias. To improve modality diversity, different data modalities including language data are crucial to capturing complex data relationships and dynamic processes.
        \item \textbf{Metadata integration and standardization}: Metadata is the crucial bridge linking diverse observation data sources, playing a key role in organizing extensive data observed from Earth's surface and atmosphere. Metadata like the data source, collection time, modality, resolution, cloud coverage, spectral bands, and preprocessing details are critical for enabling FMs with features like geolocation embedding, scale awareness, and wavelength embedding. As for weather and climate phenomena, it is important to employ standardized ontologies for each data point to maintain consistency across datasets.
\end{enumerate}

\subsection*{Model design and training}

\begin{enumerate}[a.]
\setcounter{enumi}{2}

    \item \textbf{Dynamic encoder with conditional computation}: Dynamic encoder that employs conditional computation techniques allows the model to efficiently manage a broad spectrum of inputs flexibly. Among the promising techniques to process varying input sizes are: an adaptive input projection layer designed to dynamically adjust according to the quantity of input features~\cite{xiong2024neural}; Transformers with recurrent mechanism to process sequences of variable lengths; an input chunking strategy, complemented by mechanisms that preserves context across segments. These sophisticated design strategies bolster the model’s capability to analyze complex, multi-scale data and support features like geolocation embedding, wavelength embedding, and scale awareness.
    \item \textbf{Spatio-temporal analysis across multiple scales}: EO data is captured at distinct spatial and temporal resolutions. To effectively process this complex data, it is essential to design effective spatiotemporal attention mechanisms to simultaneously capture the temporal and spatial dependencies. It facilitates an understanding of how conditions evolve and vary across different regions, thereby enhancing the model's performance in understanding Earth's dynamic systems.
    \item \textbf{Multi-modal learning framework}: An ideal FM, pre-trained with data from multiple sources, must skillfully exploit the synergy among diverse data modalities, weaving together complementary insights and common knowledge~\cite{fuller2024croma}. To accomplish this, methods that concentrate on two essential aspects of multi-modal learning need to be designed~\cite{wang2023decur,xiong2020msn}. One is the meta-modality representation learning, which extracts and synthesizes information across various modalities, creating a meta-representation that captures overarching patterns. The other is modality-specific representation learning, which utilizes modality-specific encoders to learn rich, modality-specific features.
    \item \textbf{Unified FM with geographical mixture of experts}: Designing an effective Geographical Mixture of Experts (GeoMoE) strategy is pivotal to harnessing the collective strengths of FMs in Earth and climate fields. GeoMoE integrates specialized models across various domains, enabling the framework to precisely allocate the right expertise. GeoMoE can not only enhance performance by adeptly adapting to diverse situations but also reduce computational costs by judiciously deploying resources. The adaptable nature of GeoMoE allows for the seamless integration of emerging scientific discoveries, ensuring the FMs remain at the forefront of technological and scientific progress.
    \item \textbf{Continual pre-training for carbon-minimized FMs}: Incorporating continual learning mechanisms~\cite{mendieta2023towards}, regularization techniques, and transfer learning strategies is critical to creating a carbon-minimized and adaptable Earth and climate FM. These strategies allow the model to persistently update its knowledge, enhancing performance and generalization in response to new and evolving scenarios. Techniques like elastic weight consolidation~\cite{kirkpatrick2017overcoming} to prevent catastrophic forgetting, and knowledge distillation for efficient model scaling are pivotal.
    \item \textbf{Uncertainty quantification}: In EO and climate modeling, where handling petabyte-scale datasets is a trend, efficient uncertainty quantification is critical. Sparse Gaussian processes~\cite{snelson2005sparse} offer a scalable Bayesian approach, using inducing points to manage the computational load. Ensemble methods utilize multiple models to capture a broad spectrum of outcomes, enhancing performance and providing robust uncertainty measures. Quantile regression directly computes uncertainty intervals, offering insights into prediction reliability efficiently. These candidate techniques have the potential to offer the FMs with uncertainty quantification ability required in Earth and climate sciences.
    \item \textbf{Physical consistency:} 
    Physical consistency can be achieved through the integration of physics into datasets, architectures, or loss functions, thereby augmenting the FM's generalization and interpretability.
    For instance, physical knowledge is usually represented by differential equations with analytical or numerical solutions. Physics-informed neural networks~\cite{karniadakis2021physics,xu2024large} incorporate these equations into the neural network's loss function, enabling unsupervised learning for understanding physical processes. Furthermore, employing a hybrid learning framework that integrates physical models and FMs within a differentiable neural network can result in the establishment of a structured geodynamics FM.
    \item \textbf{Alignment with large language models}: Merging EO, weather, and climate FMs with language processing capabilities is vital for enhancing their analytical depth and facilitating user interaction~\cite{liu2024visual}. By designing effective projection modules, the Earth and climate FMs can learn to align with language capabilities to gain open-vocabulary learning ability~\cite{kuo2022f}, and enrich the understanding of an extensive range of terms and concepts.
    
\end{enumerate}



\section*{Evaluating Earth Foundation Models}
Task-specific models are usually evaluated by a specific test dataset.
Summarizing model performance through numerical metrics such as accuracy or F1-score allows for easy comparison of such models.
FMs, however, are supposed to be beneficial for a wide range of tasks, posing challenges to the comparability of FMs.
In natural language processing, FMs are evaluated through so-called \emph{benchmarks}, which specify a collection of evaluation tasks.
The tasks within a benchmark are selected to evaluate different aspects of downstream applications ~\cite{paass2023knowledge}.
In computer vision, benchmarks are mostly assembled ad-hoc from existing evaluation datasets~\cite{oquab2024dinov2}.

Useful Earth FMs must learn robust representations for diverse EO , weather, and climate data.
Evaluating these models requires standardized benchmarking suites similar to the ones proposed in natural language processing to assess model performance across various tasks and datasets.
This section discusses key requirements for such benchmarks and examines existing benchmarks, highlighting their strengths and limitations.
Ultimately, we identify gaps where future benchmarks can offer even more comprehensive evaluations of Earth FMs.

\subsection*{Requirements for Earth benchmarks}
\begin{itemize}
\item\textbf{Data Diversity:}
Benchmarks for Earth FMs need to evaluate the model on diverse data.
For EO, various data modalities need to be covered, such as multispectral, SAR, and hyperspectral data.
For weather and climate data, various variables should be included, like atmospheric pressure, temperature, and precipitation.
Another aspect of data diversity is spatio-temporal coverage of the evaluation tasks, which should represent most parts of the globe and range from near real-time applications to long time-series including historical data. For climate data, even future projections should be included.
\item\textbf{Representative Tasks:}
Furthermore, the tasks in the benchmark need to be representative of downstream tasks that the models could be applied for.
For EO, this includes tasks like scene classification, object detection, segmentation, change-detection, time-series analysis, and regression.
For the weather and climate domains, possible tasks are forecasting, extreme event detection, downscaling, or emulation of physical climate models and corresponding climate scenario analysis.
\item\textbf{Interaction between components:}
Ideally, a benchmark for Earth FMs should not only evaluate the model on its individual aspects of EO, weather, and climate, but also evaluate their integration.
For this, tailored tasks shall be defined that require the interaction between components. 
Possible examples for such tasks are the downscaling weather data using satellite imagery, combining weather and satellite data to detect extreme events like droughts, or improving long-term climate projections with knowledge about short-term weather patterns.
\end{itemize}

\subsection*{Existing benchmark datasets}
\begin{table}[htbp]
    \newcommand{\rng}[3]{\SIrange[range-phrase={--}]{#1}{#2}{#3}}
    \setlength{\tabcolsep}{2pt}
    \newcommand{\rot}[1]{\multicolumn{1}{c}{\rotatebox{90}{\textbf{#1}}}}
    \begin{subtable}{\textwidth}
    \begin{tabular}{lrccccccccccrr}
        \rot{Name}&\rot{\# Tasks}&\rot{RGB}&\rot{Multispectral}&\rot{Hyperspectral}&\rot{SAR}&\rot{Time Series}&\rot{Classification}&\rot{Regression}&\rot{Segmentation}&\rot{Object Det.}&\rot{Change Det.}&\rot{Resolution}&\rot{Timespan}\\
        \midrule
        SustainBench~\cite{yeh2021sustainbench}&15&\yy&\yy&\nn&\nn&\yy&\yy&\yy&\yy&\nn&\nn&\rng{0.6}{30}{\metre}&1996--2019\\
        GEO-Bench~\cite{lacoste2024geo}&12&\yy&\yy&\yy&\yy&\nn&\yy&\nn&\yy&\nn&\nn&\rng{0.1}{15}{\metre}&2001--2021\\
        FoMo-Bench~\cite{bountos2023fomobench}&16&\yy&\yy&\yy&\yy&\nn&\yy&\nn&\yy&\yy&\nn&\rng{0.01}{60}{\metre}&2011--2023\\
        MDAS~\cite{hu2023mdas}&3&\yy&\yy&\yy&\yy&\nn&\nn&\nn&\yy&\nn&\nn&\rng{0.3}{30}{\metre}&2018\\
        PhilEO Bench~\cite{fibaek2024PhilEO}&3&\nn&\yy&\nn&\nn&\nn&\nn&\yy&\yy&\nn&\nn&\SI{10}{\metre}&N/A\\
        SkySense~\cite{guo2023skysense}&17&\yy&\yy&\nn&\yy&\yy&\yy&\nn&\yy&\yy&\yy&\rng{0.05}{30}{\metre}&2002–2019\\
        Prithvi~\cite{jakubik2023foundation}&4&\nn&\yy&\nn&\yy&\yy&\nn&\nn&\yy&\nn&\nn&\SI{30}{\metre}&2017--2022\\
        \bottomrule
    \end{tabular}
    \caption{Benchmarks for EO FMs}%
    \label{tab:benchmarks_eo}
    \end{subtable}
    \bigskip
    \begin{subtable}{\textwidth}
    \begin{tabular}{lccccccllrc}
        \rot{Name}&\rot{Nowcasting}&\rot{Medium-Range}&\rot{S2S Forecasting}&\rot{Climate Projection}&\rot{Extreme Events}&\rot{Downscaling}&\rot{Resolution (Lat)}&\rot{Resolution (Lon)}&\rot{Resolution (Time)}&\rot{Timespan}\\
        \midrule
        ERA5$^\dagger$~\cite{hersbach2020era5}
            &\yy&\yy&\yy&\yy&\nn&\nn&0.25\degree&0.25\degree&\SI{1}{\hour}&1959--2024\\
        WeatherBench~\cite{rasp2020weatherbench}
            &\nn&\yy&\nn&\nn&\nn&\nn&5.63\degree&5.63\degree&\SI{6}{\hour}&1979--2018\\
        WeatherBench 2~\cite{rasp2023weatherbench}
            &\nn&\yy&\nn&\nn&\nn&\nn&1.5\degree&1.5\degree&\SI{6}{\hour}&1979--2023\\
        ExtremeWeather~\cite{racah2017extremeweather}
            &\nn&\nn&\nn&\nn&\yy&\nn&0.23\degree&0.31\degree&\SI{6}{\hour}&1979--2005\\
        ClimateLearn~\cite{nguyen2024climatelearn}
            &\nn&\yy&\nn&\nn&\yy&\yy&5.63\degree&5.63\degree&\SI{6}{\hour}&1979--2018\\
        CMIP6$^\dagger$~\cite{o2016scenario}
            &\nn&\yy&\yy&\yy&\nn&\nn&1.25\degree&2.5\degree&\SI{24}{\hour}&1850--2100\\
        ClimateBench~\cite{watsonparris2022climatebench}
            &\nn&\nn&\nn&\yy&\nn&\nn&1.89\degree&2.5\degree&\SI{1}{\year}&1850--2100\\
        MRMS$^\dagger$~\cite{zhang2016multiradar}
            &\yy&\yy&\yy&\nn&\nn&\yy&0.01\degree&0.01\degree&\SI{2}{\minute}&2014--2024\\
    %
        \bottomrule
    \end{tabular}
    {\footnotesize $^\dagger$Raw dataset without associated evaluation protocol}
    \caption{Benchmarks for weather and climate FMs}
    \label{tab:benchmarks_climate}
    \end{subtable}
    \caption{Overview of existing benchmark datasets for EO FMs (a) and weather/climate FMs (b)}\label{tab:benchmarks}
\end{table}

For evaluating EO models,
SustainBench~\cite{yeh2021sustainbench} focuses on the United Nations' sustainable development goals,  encompassing 15 tasks.
The tasks are representative, covering classification, segmentation, regression and time-series analysis.
It evaluates on a number of optical sensors like Landsat, Sentinel-2 and MODIS. However, SAR data is not considered in this benchmark.
Finally, owing to the design of the dataset as a sustainability benchmark, the considered tasks are limited to agricultural and humanitarian applications.
More recently, GEO-Bench~\cite{lacoste2024geo} is a benchmarking suite designed specifically for evaluating EO FMs.
It includes diverse Earth-relevant tasks and has standardized evaluation protocols.
By limiting evaluation to classification and segmentation tasks, however, it omits important EO tasks, such as regression or time-series analysis.
MDAS~\cite{hu2023mdas} and PhilEOBench~\cite{fibaek2024PhilEO} are suites that trade larger datasets for a smaller number of tasks.
Finally, Prithvi~\cite{jakubik2023foundation} and SkySense~\cite{guo2023skysense} have proposed new benchmarks alongside their FMs.
Table~\ref{tab:benchmarks_eo} summarizes these benchmarks by the evaluated modalities, tasks, resolutions and timespan.


For weather data, the WeatherBench~\cite{rasp2020weatherbench} and WeatherBench~2~\cite{rasp2023weatherbench} collections provide a comprehensive benchmark for medium-range weather forecasts on a global scale.
These datasets include both historical records and model forecasts of essential meteorological variables, allowing for standardized evaluation of weather FMs.
An increasing number of weather FMs are directly evaluated on ERA5~\cite{bi2023accurate,chen2023fuxi,price2023gencast}.
ExtremeWeather~\cite{racah2017extremeweather} specifically evaluates weather models on extreme weather events.
Nowcasting is typically done only for precipitation~\cite{sonderby2020metnet}, for which highly resolved datasets such as MRMS~\cite{zhang2016multiradar} exist.
As a first multi-task benchmark, ClimateLearn~\cite{nguyen2024climatelearn} evaluates not only weather forecasting, but also extreme weather prediction and downscaling.

ClimateBench~\cite{watsonparris2022climatebench} collects physical simulations under different forcing scenarios to aid the development and evaluation of data-driven climate emulation models.
Similar to the trend of evaluating weather models directly on ERA5, some climate models also directly evaluated against CMIP6 simulation outputs~\cite{o2016scenario}.
Weather and climate benchmarks are summarized by evaluation tasks, resolution and timespan in Table~\ref{tab:benchmarks_climate}.

\subsection*{Gaps in existing benchmarks}
Currently, there is \textbf{no consensus on how to adapt a pre-trained FM for a downstream task}. Some works fine-tune the entire model, updating all parameters. Other works only train a single final network layer in a process called linear probing. Finally, parameter-efficient fine-tuning methods are another option. Benchmark results will only be comparable when models have been adapted to the downstream tasks using the same fine-tuning method. In particular, benchmarks are missing for few- and zero-shot settings.

However, \textbf{certain regions are not represented} in existing EO benchmarks.
This limits the potential for cross-disciplinary research and holistic understanding of global climate systems.
Including benchmarks for the missing regions such as the oceans~\cite{kikaki2022marida} or polar areas~\cite{gourmelon2022calving} in evaluations would facilitate better modeling of phenomena like sea-level rise or ocean circulation patterns.

Further, evaluation datasets are often filtered for specific acquisition conditions like minimal cloud cover, which is \textbf{not representative of real-world data}.
This can introduce artificial biases in the evaluation scores, making the models appear more capable than they truly are when applied to the diverse and often imperfect data encountered in actual use.
Similarly, weather and climate benchmarks are often based on reanalysis products and might benefit from evaluations on observational data as well.

For evaluating Earth FMs, the main gap in existing benchmarks is \textbf{the lack of evaluation tasks that integrate EO data with weather and climate data}.
This limits our ability to assess how well these models can capture the complex interplay between land, ocean, atmosphere, and human activity.

The rapid development of Earth FMs currently faces a challenge with \textbf{the lack of standardized evaluation benchmarks}.
With each model employing its own unique set of tasks, objective comparisons and assessments are difficult.
A convergence towards standardized models is needed to ensure fair and transparent evaluations, and allow researchers to determine the strengths and weaknesses of each model.
This, in turn, would greatly aid collaboration and accelerate the advancement of FMs for the Earth sciences.

\section*{Perspectives}

What comes after the ideal Earth FM? Identifying the most relevant research questions is crucial for maximizing the utility of the model in advancing our understanding of Earth and climate science. We give seven recommendations, as follows:

\begin{itemize}
\item\textbf{Energy-efficient Adaptation:} Upon acquiring FMs, enhancing their efficiency for downstream tasks is vital for broad applications. Future research should explore strategies to boost computational, parameter, and memory efficiencies without compromising predictive performance. Possible research directions include quantization to reduce parameter precision, pruning to eliminate redundant parameters for faster inference, and knowledge distillation to transfer insights from larger to smaller models efficiently. These efforts aim not only to broaden the models' applicability but also to advance toward more sustainable and energy-efficient ML development.

\item\textbf{Foundation model enabled sciences:}
Once an ideal Earth and climate FM becomes available, the most prominent research question is EO-informed weather and climate forecasting. As the FM is jointly trained from EO, weather and climate data, it can be explored to address the scale mismatch between observational data and the physical model of the Earth system and its climate. Exploiting  feedback at higher spatial resolution can enable new exciting forecast and management applications. One additional example is to detect and predict extremes. As the FM is learned from massive data with ``normal'' patterns, it can flag deviations with much high accuracy, helping detect ``extremes'', such as wildfires, floods, or other environmental disturbances. 

\item\textbf{Machine unlearning:}
An ideal EO FM utilizes multimodal data, such as high-resolution and multispectral images, which can violate privacy by exposing sensitive information about individuals and infrastructures. Even basic optical imagery can pose a threat to privacy. Addressing this, the concept of ``machine unlearning'', coined in 2015~\cite{cao2015towards}, aims to enable these models to effectively ``forget'' specific subsets of a dataset while minimizing negative effects on downstream tasks performances. This is crucial for compliance with regulations, as models' generalization abilities partly rely on memorization~\cite{feldman2020does}, posing a trade-off between data protection and model performance. 





\item\textbf{Continual learning:} 
Efficiently updating FMs is also important. This includes integrating data from new sensors or modalities, upgrading outdated old data with better-quality new data, and ensuring up-to-date knowledge or constraints of the model with the evolving world. A natural technique is continual learning, which aims to allow models to learn new information without forgetting the past. However, compared to continual learning on task-specific applications, continual learning with generic FMs is more challenging due to model complexity, data size, computational cost, and task-agnostic objectives. To solve these, it's crucial to develop new model architectures, training objectives, and more sophisticated optimization mechanisms.


\item\textbf{Adversarial defenses:}
Furthermore, the continuous learning of data integration process should be complemented with incorporating adversarial ML techniques to mitigate risks on FMs. This approach trains models to resist adversarial data, which are deliberately manipulated inputs designed to mislead the FM into making incorrect predictions~\cite{xu2022universal}. For instance, adversarial inputs might include subtly modified images, despite being barely noticeable, which can lead to inaccuracies in predictions and affect performance. By integrating adversarial ML, FMs become more robust and better equipped to adapt to new patterns when deployed in real-world scenarios.

\item\textbf{Interpretability:}
FMs with billions of parameters trained on massive unlabeled data present particular challenges in explaining their outputs, compared to smaller, specialized ML systems. The nature of FMs, i.e., serving as the backbone for a wide range of downstream applications, makes research on their interpretability of crucial importance. This can be achieved to a certain extent, e.g. by examining token prioritization, directly adjusting model weights, interpretable distillation, and large language model-based explanation techniques. Nevertheless, these tools may introduce biases due to inappropriate model setup, non-physical reasoning, and the absence of counterfactual reasoning capabilities and shall therefore be utilized with caution.

\item\textbf{Cross-disciplinary inspiration:}
Concepts from image FMs could inspire future improvements for Earth FMs, such as new architectures based on state-space models~\cite{zhu2024vision} or new ways of training FMs, like label-free scene understanding~\cite{chen2023labelfree}.
Following similar requirements for multimodal FMs, medical AI is currently moving towards combining highly heterogeneous sets of modalities within FMs~\cite{truhn2024large}.
For text FMs, retrieval augmented generation~\cite{lewis2020retrievalaugmented} is currently opening up many new applications. Future Earth FMs could similarly retrieve data that is geospatially or temporally close to a given query to improve their responses.
\end{itemize}

\backmatter

\section*{Declarations}

\bmhead{Acknowledgements}

This work is jointly supported by the European Commission through the project ``ThinkingEarth—Copernicus Foundation Models for a Thinking Earth'' under the Horizon 2020 Research and Innovation program (Grant Agreement No. 101130544), by the German Federal Ministry for the Environment, Nature Conservation, Nuclear Safety and Consumer Protection (BMUV) based on a resolution of the German Bundestag (grant number: 67KI32002B; Acronym: \textit{EKAPEx}), by the German Federal Ministry of Education and Research (BMBF) in the framework of the international future AI lab ``AI4EO -- Artificial Intelligence for Earth Observation: Reasoning, Uncertainties, Ethics and Beyond'' (grant number: 01DD20001), by German Federal Ministry for Economic Affairs and Climate Action in the framework of the ``national center of excellence ML4Earth'' (grant number: 50EE2201C), by the Excellence Strategy of the Federal Government and the Länder through the TUM Innovation Network EarthCare and by Munich Center for Machine Learning.

\bmhead{Author contributions}

X.Z. conceived the work and created an initial outline. All authors wrote and edited the manuscript.

\bmhead{Competing interests}

The authors declare no competing interests.

\bmhead{Data availability}

Data used to create figures in this paper are available at \url{https://github.com/zhu-xlab/EarthFoundationModels}.

\bmhead{Additional information}

Supplementary information is available for this paper.

Correspondence and requests for materials should be addressed to X.Z.

Reprints and permissions information is available at www.nature.com/reprints.

\section*{Supplementary Information}

\subsection*{Acronyms}

The following standard acronyms are used in this paper, in order of appearance:

\begin{table}[h]
\begin{tabular}{ll}
    FM & Foundation Model \\
    EO & Earth observation \\
    AI & Artificial Intelligence \\
    ML & Machine Learning \\
    ESM & Earth System Model \\
    PB & Petabytes \\
    ECMWF & European Centre for Medium-Range Weather Forecasts \\
    NOAA & National Oceanic and Atmospheric Administration \\
    SAR & Synthetic-Aperature Radar \\
    RGB & Red-Green-Blue \\
    MSI & Multispectral Imagery \\
    HSI & Hyperspectral Imagery \\
    GPU & Graphics Processing Unit \\
    TPU & Tensor Processing Unit \\
    GeoMoE & Geographic Mixture of Experts \\
    S2S & Subseasonal to Seasonal \\
\end{tabular}
\end{table}

\subsection*{The volume of big Earth data}

To plot Fig.~\ref{fig:open-eo-volume}, the accumulated volume of big Earth data, we search for available statistics about popular EO and climate archives, and fit an exponential curve for missing years and future trends. Main references include:

\begin{itemize}
    \item \textbf{Sentinel:} Copernicus Sentinel Data Access Annual Reports~\cite{annualreport}.
    \item \textbf{Landsat:} The 50-year Landsat collection 2 archive~\cite{crawford202350}, Soille \textit{et al.}~\cite{soille2018versatile}.
    \item \textbf{MODIS:} Soille \textit{et al.}~\cite{soille2018versatile}.
    \item \textbf{ERA:} The ERA5 global reanalysis~\cite{copernicus2023era5,hersbach2020era5,bell2021era5}.
    \item \textbf{CMIP:} Coupled Model Intercomparison Project Phase 6~\cite{cmip6official,petrie2020coordinating}.
    \item \textbf{Labeled datasets:} EarthNets~\cite{xiong2022earthnets}.
\end{itemize}

\subsection*{Additional details of state-of-the-art foundation models}

\subsubsection*{Earth observation foundation models}

Extended Data Table~\ref{xtab:EO-FM} lists a set of representative EO FMs, from which we selected 6 to include in the figures. Data size takes the number of pixels.
To create Fig.~\ref{fig:eo_fm}, 
\textit{modality diversity} gets the number of modalities; \textit{resolution diversity} gets the number of stripes among $\{\text{very high}, \text{high}, \text{medium}, \text{low}\}$; \textit{time series} gets the capacity level of processing time sequences among $\{\text{unable}, \text{fixed}, \text{flexible}\}$; 
\textit{spatial coverage} gets the coverage level of pretraining data among $\{\text{local}, \text{regional}, \text{global}\}$;
model size gets the log scaled number of parameters. These numbers are then scaled to 0-1.0 by dividing by the maximum for each category.

\subsubsection*{Weather and climate foundation models}

Extended Data Table~\ref{xtab:weather-models} lists the architecture, spatiotemporal resolution and range, model size, and computational requirements of several popular weather and climate FMs. Time span is the maximum attempted lead time used in each paper, and does not necessarily suggest the range over which forecasts can be made with any accuracy guarantee. Cluster size and training time are for pre-training, not inference. SEEDS is not an autoregressive model, and therefore has no temporal resolution. NeuralGCM is a family of models evaluated at different spatiotemporal resolutions. In order to create Fig.~\ref{fig:cl_fm}, resolution is inverted, the log of all numbers is taken, and all numbers are scaled to the 0.2--1.0 range.

\subsection*{Vision language foundation models for Earth observation}
\label{sec:vl_data}

\subsubsection*{Vision language model architectures}

When the FM's training dataset contains text, or semantic information that can be turned into text, existing vision-language models can easily be fine-tuned using this data, and a frozen instance of the FM. This allows for human-model interaction by the means of a large language model, and can even improve performances on downstream tasks, as empirical evidence suggests~\cite{srivastava2024omnivec}.
Vision-language models such as, for instance,  CLIP~\cite{radford2021learning}, BLIP-2~\cite{li2023blip}, and CogVLM~\cite{wang2023cogvlm} integrate visual and textual data to enhance their understanding and the human interaction capabilities they offer. CLIP processes image-text pairs by separately embedding them using modality-specific encoders and projecting them into a shared embedding space. CLIP performs "contrastive learning": when the image and the text are indeed paired (positive pair), their embeddings are brought closer through the maximization of their cosine similarity. When it is not the case (negative pair), the cosine similarity is instead minimized. CLIP offers basic human-model interaction, such as cross-modal image retrieval, and is commonly used as a text-enriched vision encoder. BLIP-2 improves over this by using an alignment module that generates embeddings of such quality that they can be directly projected into the hidden space of a large language model (LLM) along with a text prompt. This capability effectively facilitates the 'interrogation' of images, allowing for more sophisticated querying and interaction. CogVLM takes it further with a more detailed design, allowing for even better embeddings, for deeper and more meaningful human-model interactions.

In EO, RemoteCLIP~\cite{liu2023remoteclip} and SkyScriptCLIP~\cite{wang2023skyscript} showcase their ability to handle EO data without overfitting, and require far fewer parameters than off-the-shelf CLIP models pretrained on large amounts of image-text data. Similarly, RS5M~\cite{zhang2024rs5m} pushes this further by displaying high performances on simple downstream tasks (but not on complex image-text tasks yet), such as zero-shot land cover classification, thus successfully training the current largest CLIP model in EO, three times larger than RemoteCLIP and SkyScriptCLIP. However, challenges remain in training models of such size to perform well on advanced visual reasoning tasks within remote sensing, highlighting an open problem related to the scarcity of textual data and its mixed quality in this domain. More recent initiatives, relying on much larger models, such as EarthGPT~\cite{zhang2024earthgpt} developed for multisensor RS data comprehension, exhibits high performance across different tasks compared to some specialist models.

\subsubsection*{Curated vision-language datasets}

Advances in vision-language models have sparked interest in the remote sensing community, owing to their promising results in visual reasoning. As argued above, there is a need for datasets that combine visual data with semantically rich text annotations. Extended Data Table~\ref{xtab:language-models} shows vision-language datasets for EO. The most explored image-text task in EO is image captioning~\cite{8240966, 7546397, 9866055}, requiring the creation of sentences that describe the corresponding image. More complex tasks include visual question answering~\cite{9444570, lobry2020rsvqa, 9553307} and visual grounding~\cite{10.1145/3503161.3548316, 10056343}, which involve answering questions based on images and identifying specific objects within those images, respectively.

Complex, exhaustive text data is crucial for training vision-language models, highlighting the importance of rich, detailed image-text combinations to improve their visual reasoning skills. Despite the potential benefits, the availability of such datasets is limited, primarily due to the need for manual or costly Multimodal Large Language Model annotating, the exclusive ownership of some data by specific entities, the demonstrated impossibility to collect remote sensing image-text pairs at scale using web-crawled large-scale datasets~\cite{NEURIPS2022_a1859deb}, and model filtering~\cite{wang2023skyscript}. To remedy to these issues, efforts have been made to generate remote sensing image-text pairs through indirect methods, such as leveraging semantic segmentation and object detection datasets to derive semantic information in the form of text~\cite{liu2023remoteclip, 10056343, 9444570, 9553307, 10056343}, cross-referencing OpenStreetMap geotags with Google Earth Engine images to craft comprehensive captions or question/answer pairs~\cite{wang2023skyscript, 10.1145/3503161.3548316, lobry2020rsvqa}. However, these methods yield a very limited fraction of usable data compared to the vast collections available for natural images, such as the extensive LAION-5B dataset~\cite{NEURIPS2022_a1859deb}, on top of often producing non-natural descriptions. Recent initiatives like ChatEarthNet~\cite{yuan2024chatearthnet} or RS5M~\cite{zhang2024rs5m} demonstrate the ongoing efforts to construct EO image-text datasets at scale by leveraging the capabilities of advanced vision-language models, such as GPT4-V~\cite{gpt-4vision} that performs well in image captioning, but that is costly. 

\subsection*{Distribution perspectives of foundation models}

The majority of users who could potentially benefit from a FM are domain experts (as opposed to ML researchers) and non-technical policy experts with limited ML experience. It is paramount that a unified Earth and climate FM is as easy to use as ChatGPT and SAM, and does not require a Ph.D. in computer science to use. The base model architecture and weights can be distributed through existing popular geospatial ML libraries like Raster Vision~\cite{Azavea_Element_84_Raster_Vision_An} or TorchGeo~\cite{Stewart_TorchGeo_Deep_Learning_2022}, with an additional graphical user interface for non-technical users. These libraries are already well tested across a variety of Python versions and platforms and have large dedicated userbases. An interactive interface that allows users to create labels for new benchmark datasets similar to SAM could greatly speed up research on a number of important topics.

Another concern is the license under which the model weights are released. Many potential EO data sources like Planet and Maxar data may prevent the public release of pretrained model weights and cannot be used to train the model. Ideally, the weights should be released under a permissive open source license like MIT, Apache, or BSD so as to benefit the greatest number of people. However, such a FM can also be used for harm, including applications like illegal surveillance, military aggression, and resource exploitation~\cite{slonecker1998emerging}. Given these concerns, it may make sense to consider ethical software licenses like the Hippocratic License that prohibit such activities, although this can be challenging if not impossible to legally enforce.

\subsection*{Attributions}
Fig.~\ref{fig:ECFM}: Background modified from template by PresentationGO (www.presentationgo.com). Icons from Adobe Stocks (stock.adobe.com):  126670692, by KundraK; 764196510, by AbtoCreative; 509096452, by Barudak Lier; 444274525, by  vectorsanta; 734797956, SkyLine.

Fig.~\ref{fig:perspective_FM}: Icons from Adobe Stocks: 398214492, by Vectorina; 589636564, by Gopal; 489357227, by SilenceVideo; 622171957, by lovemask; 734797956, by SkyLine; 265484642, by alekseyvanin; 583742888, by barks; 534436638, AmethystStudio; 593497565, LeonART. Sentinel-1 image from ESA (©ESA). Sentinel-2, -5p image from Wikipedia. 

\section*{Extended Data}

\renewcommand\figurename{Extended Data Fig.}
\setcounter{figure}{0}

\renewcommand\tablename{Extended Data Table}
\setcounter{table}{0}

\begin{sidewaystable}[htbp]
    \centering
    \caption{State-of-the-art EO FMs. The slash sign \textit{/} indicates separate models for each; the range sign \textit{-} indicates a rough range as the exact numbers are not available; the comma sign \textit{,} indicates a combined coverage.}
    \label{xtab:EO-FM}
    \footnotesize
    \begin{tabular}{@{}llrcllrlc@{}}
        \toprule
        \textbf{Model} & \shortstack{\textbf{Pre-training} \\ \textbf{Dataset}} & \shortstack{\textbf{Data Size} \\ \textbf{(\# pixels)}} & \textbf{Modality} & \shortstack{\textbf{Spatial} \\ \textbf{Resolution}} & \shortstack{\textbf{Spatial} \\ \textbf{Coverage (km\textsuperscript{2})}} & \shortstack{\textbf{Model Size} \\ \textbf{(\# params)}} & \shortstack{\textbf{Temporal} \\ \textbf{Coverage}} & 
        \shortstack{\textbf{Sequence} \\ \textbf{Length}}
        \\ \midrule
        SeCo~\cite{manas2021seasonal} & SeCo & 70B & MS & 10m & Global (1.4M) & 23M & N/A & 1 \\
        SatMAE~\cite{cong2022satmae} & fMoW & 50B & MS / RGB & 0.5m / 10m & Global (1.7M) & 305M & 5 years & 3 \\
        RVSA~\cite{wang2022advancing} & MillionAID & 4B & RGB & 0.5m--150m & Global (18K) & 89M & N/A & 1 \\
        Scale-MAE~\cite{reed2023scale} & fMoW-RGB & 80B & RGB & 0.3--10m & Global (1.7M) & 305M & N/A & 1 \\
        CROMA~\cite{fuller2024croma} & SSL4EO-S12 & 140B & MS, SAR & 10m & Global (1.8M) & 305M & N/A & 1 \\
        DeCUR~\cite{wang2023decur} & \makecell[l]{SSL4EO-S12\\/ GeoNRW} & 140B & MS, SAR / RGB,DEM & 10m / 1m & Global (1.8M) & 23M & N/A & 1 \\
        Presto~\cite{tseng2023lightweight} & Presto & 4B & \makecell[l]{MS, SAR,\\ERA5, DEM, etc.} & 10m & Global (2K) & 400K & 2 years & 24 \\
        Prithvi~\cite{jakubik2023foundation} & HLS & 158B & MS & 10m, 30m & US & 100M & 1+ years & 3 \\
        SpectralGPT~\cite{hong2023spectralgpt} & fMoW+BigEarthNet & 12B & MS & 10m & Global (1.7M) & 86M & N/A & 1 \\
        SkySense~\cite{guo2023skysense} & SkySense & 97T & MS, RGB, SAR & 0.3m, 10m & Global (8.8M) & 2B & ? & 20 \\
        OFA-Net~\cite{xiong2024one} & \makecell[l]{Satlas.+5B pix.\\+HyspecNet.} & 13B & MS, RGB, SAR, HS & 1m, 4m, 10m, 30m & Global (260K) & 86M & N/A & 1 \\
        DOFA~\cite{xiong2024neural} & \makecell[l]{Satlas.+5B pix.\\+HyspecNet.} & 105B & MS, RGB, SAR, HS & 1m, 4m, 10m, 30m & Global (2.6M) & 305M & N/A & 1 \\
        SSL4EO-S12~\cite{wang2023ssl4eo} & SSL4EO-S12 & 140B & MS, SAR & 10m & Global (1.8M) & 23M & N/A & 1 \\
        SSL4EO-L~\cite{stewart2024ssl4eo} & SSL4EO-L & 348B & MS & 30m & Global (15.6M) & 23M & N/A & 1 \\
        SatlasPretrain~\cite{bastani2023satlaspretrain} & SatlasPretrain & 17T & MS, RGB, SAR & 0.5-2m, 10m & Global (21M) & 88M & 11 years & 10 \\
        SkyScriptCLIP~\cite{wang2023skyscript} & SkyScript & 130B & MS, RGB, text & 0.5m, 1--2m, 10-30m & Global (5M) & 305M & N/A & 1 \\ \bottomrule
    \end{tabular}
\end{sidewaystable}

\begin{sidewaystable}[htbp]
    \centering
    \caption{State-of-the-art weather and climate FMs.}
    \label{xtab:weather-models}
    \begin{tabular}{@{}lllcccccccc@{}}
        \toprule
        \textbf{Model} & \textbf{Affiliation} & \shortstack{\textbf{Model}\\\textbf{Architecture}} & \shortstack{\textbf{Temporal}\\\textbf{Resolution}} & \shortstack{\textbf{Time}\\\textbf{Span}} & \shortstack{\textbf{Spatial}\\\textbf{Resolution}} & \shortstack{\textbf{Spatial}\\\textbf{Coverage}} & \shortstack{\textbf{Model Size}\\\textbf{(\# params)}} & \shortstack{\textbf{Cluster}\\\textbf{Size}} & \shortstack{\textbf{Training}\\\textbf{Time}} \\
        \midrule
        KeislerNet~\cite{keisler2022forecasting} & Descartes & GNN & 6 hrs & 6 days & 1\degree{} & Global & 6.7M & 1 A00 & 5.5 days \\
        FourCastNet~\cite{bonev2023spherical} & NVIDIA & Transformer/FNO & 6 hrs & 200 hrs & 0.25\degree{} & Global & 433M--2.19B & 64 A100 & 16 hrs \\
        Pangu-Weather~\cite{bi2023accurate} & Huawei & Transformer & 1 hr & 7 days & 0.25\degree{} & Global & 256M & 192 V100 & 15 days \\
        GraphCast~\cite{lam2023learning} & Google & GNN & 6 hrs & 10 days & 0.25\degree{} & Global & 36.7M & 32 TPU v4 & 4 wks \\
        ClimaX~\cite{nguyen2023climax} & UCLA & Transformer & 6 hrs & 1 month & 0.25\degree{} & Global & 108--111M & 80 V100 & 3 days \\
        FengWu~\cite{chen2023fengwu} & Shanghai & Transformer & 6 hrs & 14 days & 0.25\degree{} & Global & 427M & 32 A100 & 17 days \\
        W-MAE~\cite{man2023w} & Chengdu & Transformer & 6 hrs & 9 days & 0.25\degree{} & Global & 96.5M & 8 A800 & 223.4 hrs \\
        FuXi~\cite{chen2023fuxi} & Fudan & Transformer & 6 hrs & 15 days & 0.25\degree{} & Global & 1.56B & 8 A100 & 30 hrs \\
        SEEDS~\cite{li2023seeds} & Google & Diffusion & N/A & 16 days & 2\degree{} & Global & 114M & 16 TPU v4 & 18 hrs \\
        NeuralGCM~\cite{kochkov2023neural} & Google & Differential GCM & 3.75--12 mins & 15 days & 0.7--2.8\degree{} & Global & 115--311M & 128 TPU v5e & 10 days \\
        Stormer~\cite{nguyen2023scaling} & UCLA & Transformer & 6 hrs & 14 days & 1.4\degree{} & Global & 100--600M & 128 A100 & 24 hrs \\
        GenCast~\cite{price2023gencast} & Google & Diffusion & 12 hrs & 15 days & 1\degree{} & Global & 60M & 32 TPU v4 & 5 days \\
        FengWu-GHR~\cite{han2024fengwu} & Shanghai & Transformer & 6 hrs & 10 days & 0.09\degree{} & Global & 4B & 16 A100 & 3 days \\ 
        \bottomrule
    \end{tabular}
\end{sidewaystable}

\begin{sidewaystable}[htbp]
\centering
    \caption{Vision-language datasets in EO}
    \label{xtab:language-models}
    \begin{tabular}{@{}lcrrrlrrl@{}}
        \toprule
        \textbf{Dataset} & \textbf{Year} & \textbf{\# Images} & \textbf{\# Sentences} & \textbf{Annotation} & \textbf{Image Size} & \textbf{Resolution} & \textbf{Source}\\
        \midrule
        UCM-Captions~\cite{7546397} & 2016 & 2100 & 10500 & Manual & 256$\times$256 & \SI{0.3}{\metre} & USGS\\
        Sydney-Captions~\cite{7546397} & 2016 & 613 & 3065 & Manual & 500$\times$500 & \SI{0.6}{\metre} & Google Earth\\
        RSICD~\cite{8240966} & 2017 & 10921 & 54605 & Manual & 224$\times$224 & \SIrange{0.15}{15}{\metre} & Google Earth, Baidu\\
        NWPU-Captions~\cite{9866055} & 2022 & 31500 & 157500 & Manual & 500$\times$500 & \SIrange{0.2}{0.3}{\metre} & Google Earth\\
        RSITMD~\cite{yuan2022exploring} & 2022 & 4743 & 23715 & Manual & 224$\times$224 & various & Google Earth, RSICD\\
        RSICap~\cite{hu2023rsgpt} & 2023 & 2585 & ? & Manual & 512$\times$512 & various & GF-2, JL-1\\
        RS5M~\cite{zhang2024rs5m} & 2023 & 5M & 25M & Manual/Auto & 512$\times$512 & various & Google Earth, GF-2\\
        SkyScript~\cite{wang2023skyscript} & 2023 & 2.6M & 2.6M & OSM (auto) & various & \SIrange{0.1}{40}{\metre} & Google Earth Engine\\
        ChatEarthNet~\cite{yuan2024chatearthnet} & 2024 & 163,488 & 163,484 & ChatGPT (auto) & 256$\times$256 & \SI{10}{\metre} & Sentinel-2\\
        LHRS-Align~\cite{muhtar2024lhrs} & 2024 & 1.15M & 1.15M & Existing + GE (manual/auto) & 768$\times$768 & \SI{1}{\metre} & various\\
        LAION-EO~\cite{czerkawski2023laion} & 2023 & 24,933 & 24,933 & CLIP (auto) & Avg. 633$\times$843 & various & Internet\\
        RSVQA-LR~\cite{lobry2020rsvqa} & 2020 & 772 & 77,232 & OSM (manual/auto) & 256$\times$256 & \SI{10}{\metre} & Sentinel-2\\
        RSVQA-HR~\cite{lobry2020rsvqa} & 2020 & 10,659 & 1.07M & OSM (manual/auto) & 512$\times$512 & \SI{0.15}{\metre} & USGS HRO\\
        RSVQA x BEN~\cite{9553307} & 2021 & 590,326 & 14.76M & Land cover & cf. BEN~\cite{sumbul2019bigearthnet} & cf. BEN & cf. BEN \\
        RSIVQA~\cite{9444570} & 2022 & 37,264 & 111,134  & Manual/Auto & various & \SIrange{0.1}{1}{\metre} & various\\
        FloodNet VQA~\cite{rahnemoonfar2021floodnet} & 2020 & 3,200 & 11,000 & Manual (UAV) & 224$\times$224 / 3000$\times$4000 & UAV & UAV\\
        CD-VQA~\cite{yuan2022change} & 2022 & 2968 pairs & 122000 Q\&A & Automatic & 512x512 & \SIrange{0.5}{3}{\metre} & Aerial platforms and sensors\\
        RSIEval~\cite{hu2023rsgpt} & 2023 & 100 & 936 & Manual & 512$\times$512 & HR & Google Earth, GF-2, JL-1\\
        LEVIR-CC~\cite{9934924} & 2023 & 10077 pairs & 50385 & Manual & 256$\times$256 & \SI{0.5}{\metre} & Google Earth\\
        DIOR-RSVG~\cite{10056343} & 2023 & 17402 & 38320 & Automatic & 800$\times$800 & \SIrange{0.5}{30}{\metre} & Varied\\
        VGRSI~\cite{10.1145/3503161.3548316} & 2022 & 4239 & 7933 & Automatic & 1024$\times$1024 & Various & Google Earth, various sensors\\
        MMRS-1M~\cite{zhang2024earthgpt} & 2024 & 1005842 & 1005842 & Automatic & Various & Various & Varied, includes SAR/infrared\\
        \bottomrule
    \end{tabular}
\end{sidewaystable}

\begin{figure}[htbp]
    \centering
    \includegraphics[scale=0.31]{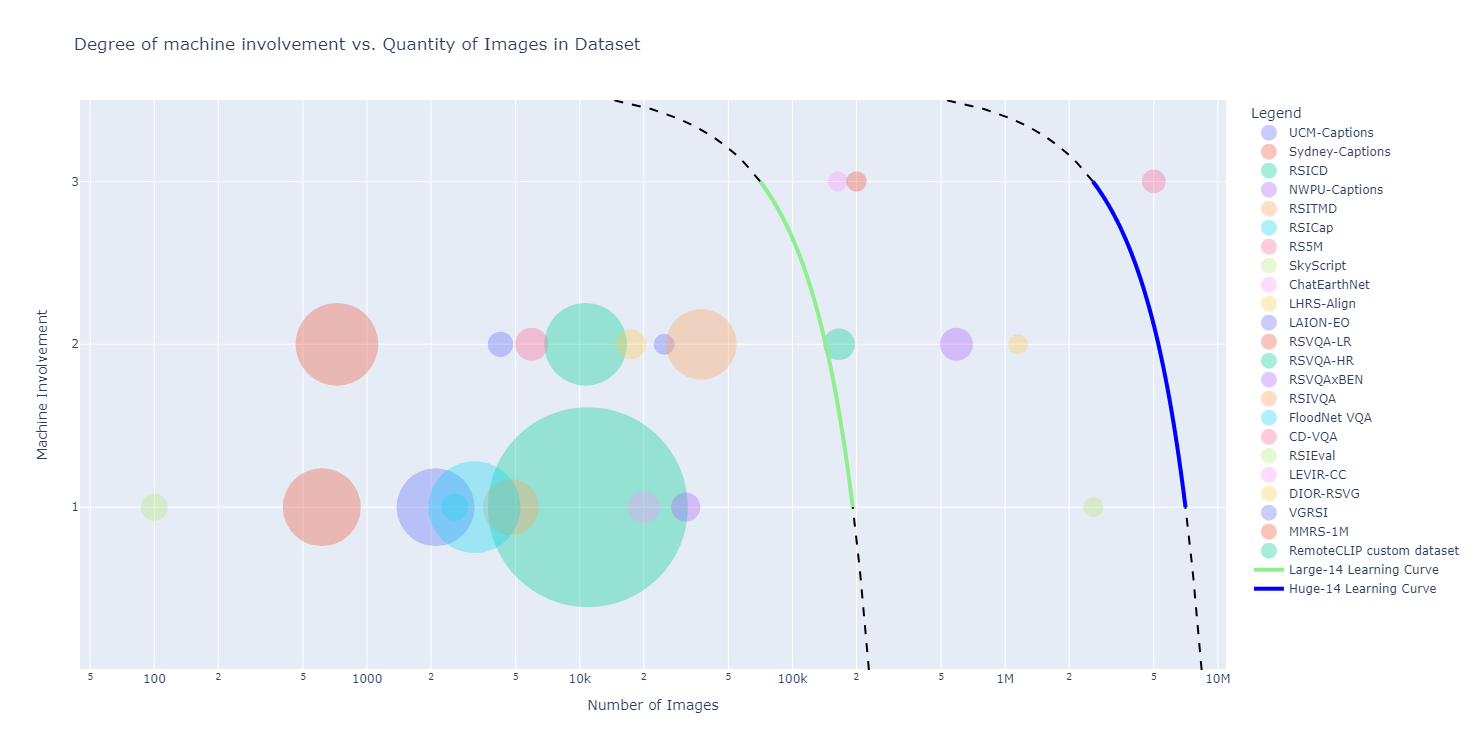}
    \caption{Visual reasoning datasets from the fields of Visual Question Answering, Remote Sensing Image Captioning, Change Detection Visual Question Answering, and Image-Text Cross-Modal Retrieval in remote sensing, with data point sizes proportional to citation counts. The curves represent CLIP models' learning and overfitting "frontiers": CLIP-ViT-L-14 shows a transition from overfitting to learning for all sorts of image-text downstream tasks, as shown by SkyScript and RemoteCLIP~\cite{wang2023skyscript, liu2023remoteclip} and illustrated by the green curve. On the far right, CLIP-ViT-H-14 has the potential for learning adequate multimodal representations across simple downstream tasks without overfitting, but is still prone to poorer performances than smaller models as the downstream tasks get more complex, as demonstrated empirically by RS5M~\cite{zhang2024rs5m}. This is indicative of overfitting. The \(y\)-axis represents the extent to which the dataset creation relied on automatic methods. Manual annotations are given a note of 1, machine-assisted human annotations and fully automatized annotations are respectively given a 2 and a 3. Note that the datasets from the figure might overlap.}
    \label{xfig:vl}
\end{figure}

\FloatBarrier

\bibliography{sn-bibliography}

\end{document}